\documentclass{IOS-Book-Article}

\usepackage{mathptmx}

\def\hb{\hbox to 10.7 cm{}}

\usepackage{latexsym}
\usepackage{epstopdf}
\usepackage[utf8]{inputenc}
\usepackage{todonotes}
\usepackage{amsmath}
\usepackage[linesnumbered,noend, algo2e]{algorithm2e}
\usepackage{csquotes}
\usepackage{textcomp}
\usepackage{subcaption}

\DeclareMathOperator*{\argmax}{arg\,max}

\usepackage{mathptmx}      % use Times fonts if available on your TeX system
\begin{document}

\pagestyle{headings}
\def\thepage{}

\begin{frontmatter}              % The preamble begins here.

%\pretitle{Pretitle}
\title{Joint learning of ontology and semantic parser from text}

\markboth{}{November 2015\hb}
%\subtitle{Subtitle}

\author[A]{\fnms{Janez} \snm{Starc}%
\thanks{Corresponding Author: Janez Starc, Jožef Stefan International Postgraduate School, Jamova 39,
1000 Ljubljana, Slovenia; E-mail:
janez.starc@ijs.si}},
\author[A]{\fnms{Dunja} \snm{Mladenić}}

\runningauthor{J. Starc et al.}
\address[A]{Jožef Stefan International Postgraduate School, Slovenia}

\begin{abstract}
Semantic parsing methods are used for capturing and representing semantic meaning of text. Meaning representation  capturing all the concepts in the text may not always be available or may not be sufficiently complete. Ontologies provide a structured and reasoning-capable way to model the content of a collection of texts.
In this work, we present a novel approach to joint learning of ontology and semantic parser from text.
The method is based on semi-automatic induction of a context-free grammar from semantically annotated text.
The grammar parses the text into semantic trees. Both, the grammar and the semantic trees are used to learn the ontology on several levels -- classes, instances, taxonomic and non-taxonomic relations. 
The approach was evaluated on the first sentences of Wikipedia pages describing people.
\end{abstract}

\end{frontmatter}
\markboth{November 2015\hb}{November 2015\hb}

	\section{Introduction}
	\label{sec:intro}

	One of the ultimate goals of Natural Language Processing (NLP) is machine reading \cite{etzioni2006machine}, the automatic, unsupervised understanding of text. 
	One way of pursuing machine reading is by semantic parsing, which transforms text into its meaning representation. However, capturing the meaning is not the final goal, the meaning representation needs to be predefined and structured in a way that supports reasoning. 
	Ontologies provide a common vocabulary for meaning representations and support reasoning, which is vital for understanding the text. To enable flexibility when encountering new concepts and relations in text, in machine reading we want to be able to learn and extend the ontology while reading.    
	Traditional methods for ontology learning \cite{cimiano2005text2onto, Zouaq:2009:EGD:1638612.1639245} are only concerned with discovering the salient concepts from text. Thus, they work in a macro-reading fashion \cite{Mitchell:2009:PSW:1693684.1693754}, where the goal is to extract facts from a large collection of texts, but not necessarily all of them, as opposed to a micro-reading fashion, where the goal is to extract every fact from the input text. Semantic parsers operate in a micro-reading fashion. Consequently, the ontologies with only the salient concepts are not enough for semantic parsing. 
	Furthermore, the traditional methods learn an ontology for a particular domain, where the text is used just as a tool. On the other hand, ontologies are used just as tool to represent meaning in the semantic parsing setting.       
	When developing a semantic parser it is not trivial to get the best meaning representation for the observed text, especially if the content is not known yet. Semantic parsing datasets have been created by either selecting texts that can be expressed with a given meaning representation, like Free917 dataset \cite{2013-acl-textual-schema-matching}, or by manually deriving the meaning representation given the text, like Atis dataset \cite{Dahl:1994:ESA:1075812.1075823}. In both datasets, each unit of text has its corresponding meaning representation. While Free917 uses Freebase \cite{Bollacker:2008:FCC:1376616.1376746}, which is a very big multi-domain ontology, it is not possible to represent an arbitrary sentence with Freebase or any other existing ontology.
	
\begin{center}
		\begin{figure}[ht]
			\centering
			\includegraphics[width=1.0\textwidth]{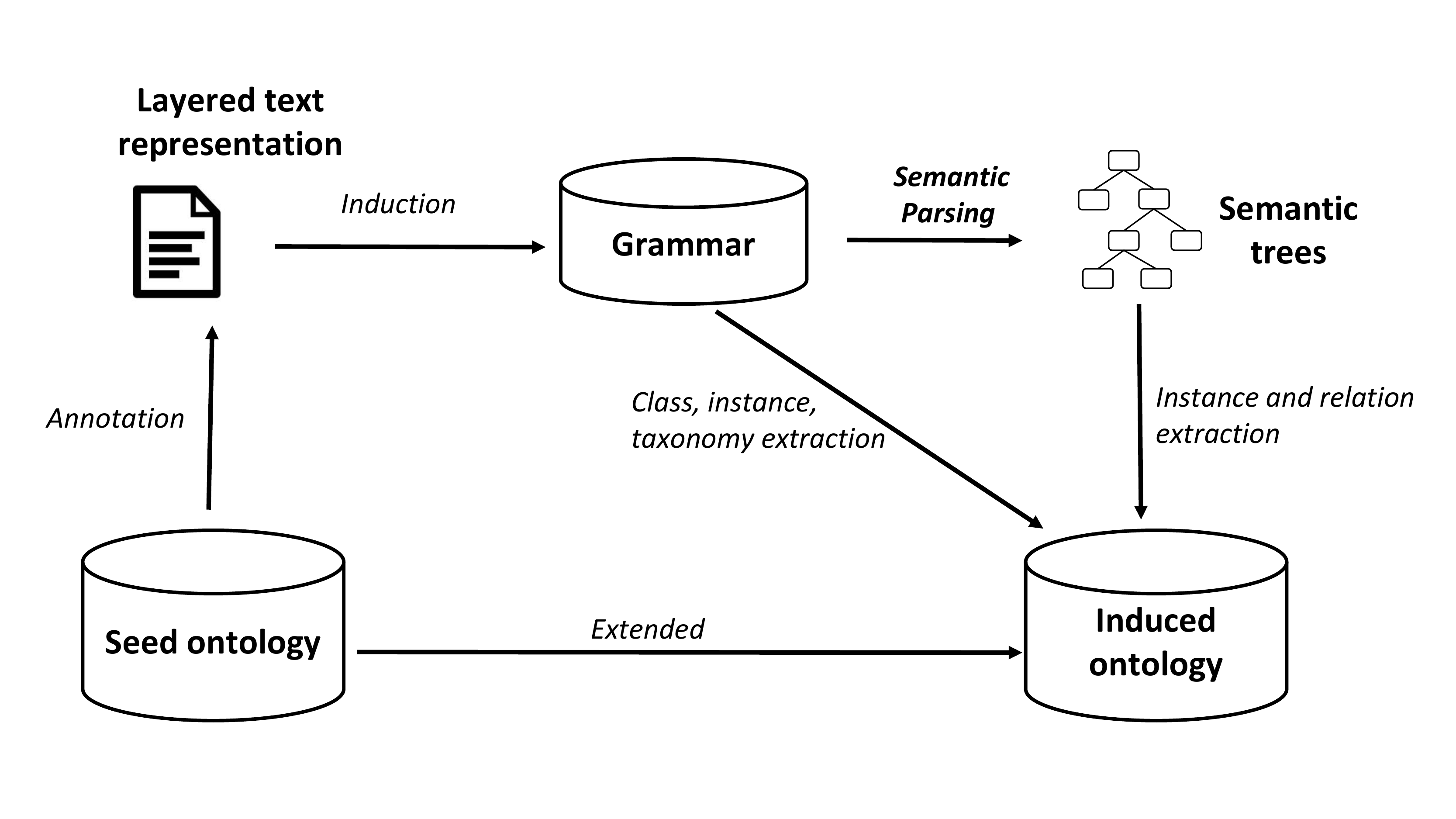}
			\caption{\small{The proposed approach to ontology and grammar induction gets a collection of text and the seed ontology as input and outputs a new ontology. The collection of text is first annotated at different levels including annotations with the concepts from existing ontology. The annotated text is then used for grammar induction, where the text is represented as semantic trees. These are used together with the grammar to induce a new ontology.}}
			\label{fig:architecture}
		\end{figure}
	\end{center}

	In this paper, we propose a novel approach to joint learning of ontology and semantic parsing, which is designed for homogeneous collections of text, where each fact is usually stated only once, therefore we cannot rely on data redundancy. Our approach is text-driven, semi-automatic and based on grammar induction. It is presented in Figure~\ref{fig:architecture}.The input is a seed ontology together with text annotated with concepts from the seed ontology. The result of the process is an ontology with extended instances, classes, taxonomic and non-taxonomic relations, and a semantic parser, which transform basic units of text, i.e sentences, into \emph{semantic trees}.
	Compared to trees that structure sentences based on syntactic information, nodes of semantic trees contain semantic classes, like location, profession, color, etc. Our approach does not rely on any syntactic analysis of text, like part-of-speech tagging or dependency parsing.
	The grammar induction method works on the premise of curriculum learning \cite{bengio2009curriculum}, where the parser first learns to parse simple sentences, then proceeds to learn more complex ones. The induction method is iterative, semi-automatic and based on frequent patterns. A context-free grammar (CFG) is induced from the text, which is represented by several layers of semantic annotations. 
	The motivation to use CFG is that it is very suitable for the proposed alternating usage of top-down and bottom-up parsing, where new rules are induced from previously unparsable parts. Furthermore, it has been shown by \cite{Kallmeyer:2010:PBC:1895073} that CFGs are expressive enough to model almost every language phenomena.  The induction is based on a greedy iterative procedure that involves minor human involvement, which is needed for seed rule definition and rule categorization. Our experiments show that although the grammar is ambiguous, it is scalable enough to parse a large dataset of sentences.
	
	The grammar and semantic trees serve as an input for the new ontology. Classes, instances and taxonomic relations are constructed from the grammar. We also propose a method for discovering less frequent instances and their classes, and a supervised method to learn relations between instances. Both methods work on semantic trees. 
	
	For experimentation, first sentences of Wikipedia pages describing people are taken as a dataset. These sentences are already annotated with links to other pages, which are also instances of DBpedia knowledge base \cite{dbpedia-swj}. Using relations from DBpedia as a training set, several models to predict relations have been trained and evaluated.  
	
	The rest of the paper is organized in the following way. 
	The grammar induction approach is presented in Section~\ref{sec:approach}. 
	%First, the layered structure of the textual data is described in Section~\ref{sec:datasets}, then the grammar is defined in Section~\ref{sec:grammar}, the description of the top-down parser follows in Section~\ref{sec:parser}, the rule induction step is presented in Section~\ref{sec:rule_induction}, the section concludes with the description of seed rules in Section~\ref{sec:seed}.
	The ontology induction approach follows in Section~\ref{sec:onto_induction}. 
	%Induction from grammar is presented in Section~\ref{sec:from_grammar} and relation extraction from semantic trees in Section~\ref{sec:tree_rule_learning}.
	In Section~\ref{sec:experiments} we present the conducted experiments with grammar induction, and instance and relation extraction. We examine the related work in Section~\ref{sec:related}, and conclude with the discussion in Section~\ref{sec:discussion}.
	
	\section{Grammar induction}
	\label{sec:approach}
	
	In this section, we propose a semi-automatic bootstrapping procedure for grammar induction, which searches for the most frequent patterns and constructs new production rules from them. 
	One of the main challenges is to make the induction in a way that minimizes human involvement and maximizes the quality of semantic trees. 
	
\begin{center}
		\begin{figure}[ht]
			\centering
			\includegraphics[width=1.0\textwidth]{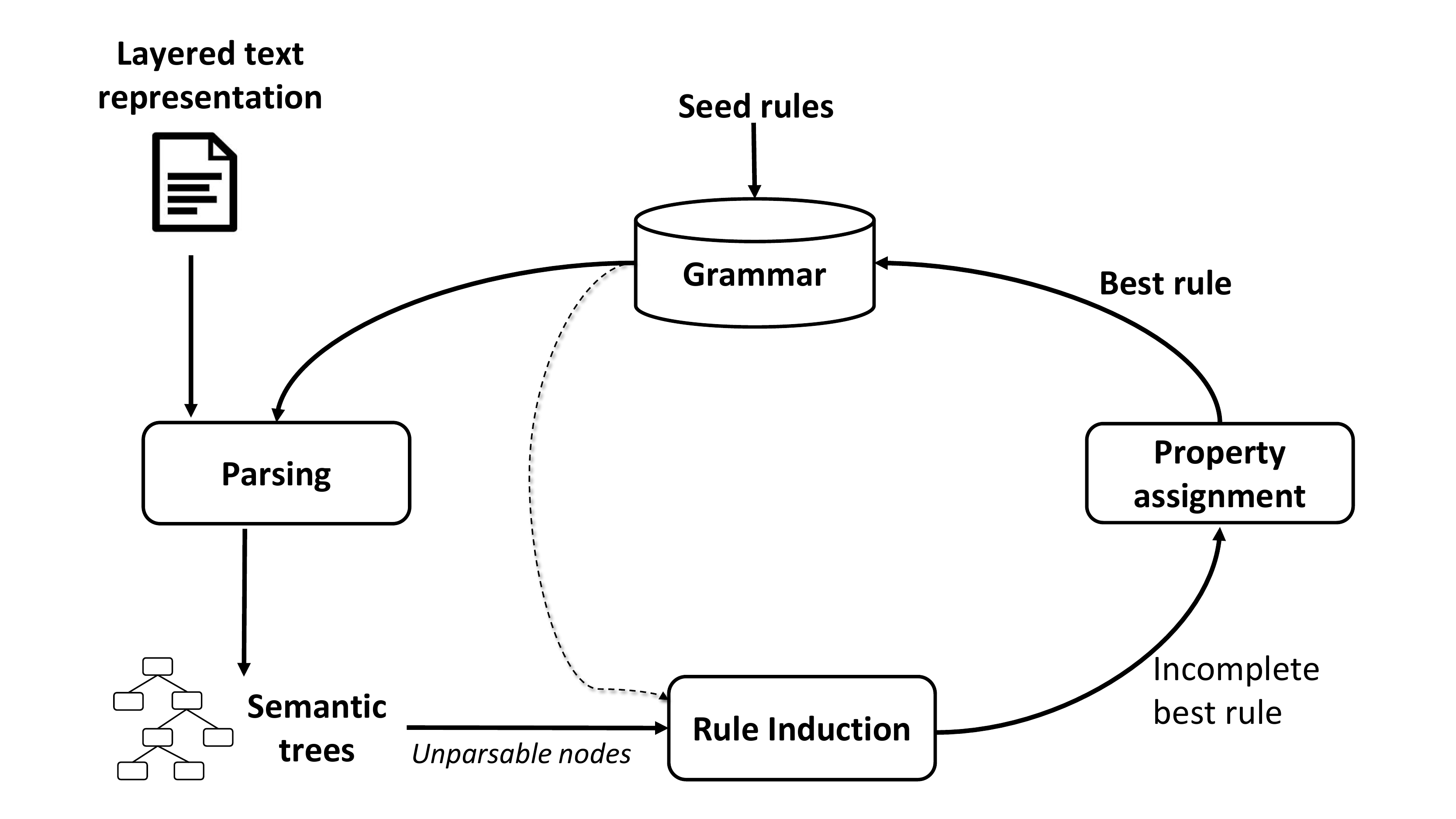}

			\caption{Grammar induction}
			\label{fig:ginduction}
		\end{figure}
	\end{center}

	The input to the process, which is illustrated in Figure~\ref{fig:ginduction}, is a set of predefined seed grammar rules (see Section~\ref{sec:seed}) and a sample of sentences in a layered representation (see Section~\ref{sec:textdata}) from the dataset. 
	The output of the process is a larger set of rules forming the induced grammar.  
	One rule is added to the grammar on each iteration. At the beginning of each iteration all the sentences are parsed with a top-down parser.
	The output of parsing a single sentence is a semantic tree -- a set of semantic nodes connected into a tree. 
	Here we distinguish two possible outcomes of the parsing: 1) the sentence was completely parsed, which is the final goal and 2) there is at least one part of the sentence that cannot be parsed. 
	From the perspective of a parser the second scenario happens when there is a node that cannot be parsed by any of the rules. We name these nodes -- \emph{null nodes} -- and they serve as the input for the next step, the rule induction. 
	In this step several rules are constructed by generalization of null nodes. The generalization (see Section~\ref{sec:rule_induction}) is based on utilization of semantic annotations and bottom-up composition of the existing rules. 
	Out of the induced rules, a rule with the highest frequency (the one that was generalized from the highest number of null nodes) is added to the grammar. 
	To improve quality of the grammar, the rules are marked by so called property, which instructs the parser how to use the rule (eg., us it in parsing but not in induction). 
	The property vitally affects result of the parsing in the following iterations potentially causing a huge semantic drift for the rest of process. 
	Consequently, a human user needs to mark the property of each rule. 
	The iterative process runs until a predefined stopping criteria is met. The criteria is either connected to the quality of the grammar or time limitation. 
	
	For the sake of transparency of the experiments, the human is involved in the beginning, when the seed rules are created and later when the rule properties are specified. However, in another setting the user could also define new rules in the middle of the bootstrapping procedure.
	
	In the following sections, we describe each component of the process in more details.
	
	\subsection{Textual data representation}
	\label{sec:textdata}
	
	The input textual data needs to be properly structured in order to work best with the proposed algorithms. Shallow NLP tools, like sentence splitting, word tokenization, named entity recognition, might help obtaining this structure. 
	The basic unit is a sentence, represented by several layers. An example is presented in Table~\ref{table:layers}. 
	Each layer consists of several tokens, which span over one or more words. 
	The basic layer is the lexical layer, where each token represents a single word. 
	All other layers are created from the annotations. 
	Some annotations, like named-entities, may span over several words; some of the words may not have an annotation, thus they are given a \emph{null token}. 
	It is crucial that all algorithms are aware how to deal with a particular layer. 
	For instance, the parser must not break apart a multi-word annotation. 
	Some layers may be derived from others using the seed ontology. 
	For example, \emph{instance} layer contains annotations to instances of the ontology and the derived \emph{class} layer represents the classes of these annotations, which are also from the ontology. 
	Annotation layers are valuable if they provide good means for generalization or connection with the ontology.  
	A \emph{term} is a subpart of the sentence, defined by the starting and ending position in the sentence. It has different interpretation in each layer. 
	If the interpretation breaks any of the tokens, it is not valid. 
	For instance, term representing \emph{Madeira} is not valid in \emph{named-entity} layer in Table \ref{table:layers} because it breaks \emph{Person}.   
	
	\begin{table}
		\centering

		\begin{tabular}{|l|c|c|c|c|c|c|c|}
			\hline
			\textbf{Layer} & \multicolumn{7}{|c|}{\textbf{Tokens}} \\ \hline
			lexical & Phil & Madeira & is & a & musician & from & Nashville \\ \hline
			small-caps &  phil & madeira & is & a & musician & from & nashville \\ \hline
			named-entity & \multicolumn{2}{|c|}{Person} & 
			\emph{-} & \emph{-} & \emph{-} & \emph{-} & Location  \\ \hline
			
			instance & \multicolumn{2}{|c|}{Phil\_Madeira} & 
			\emph{-} & \emph{-} & 
			\begin{tabular}{@{}c@{}}Musical\_ \\ Artist \end{tabular} & 
			\emph{-} & 
			\begin{tabular}{@{}c@{}}Nashville\_ \\ Tennessee \end{tabular}  \\ \hline
			
			class & \multicolumn{2}{|c|}{Person} & \emph{-} & \emph{-} & 
			Profession & \emph{-} & Location  \\ \hline
		\end{tabular} 
		\caption{Layered representation of a sentence. Null tokens are expressed as ''\emph{-}''.} 
		\label{table:layers}    
	\end{table}

	\subsection{Grammar Definition}
	\label{sec:grammar}
	
	Our context-free grammar $G$ is defined by the 5-tuple: $G = (V, \sigma, P, S, R)$, where
	\begin{itemize}
		\item $V$ is a set of \emph{non-terminals}. Each non-terminal represents a semantic class, e.g. $\langle \text{Person} \rangle$,  $\langle \text{Color} \rangle$, $\langle \text{Organization} \rangle$. There is also a universal non-terminal  $\langle * \rangle$, which can be replaced by any other non-terminal. The same non-terminal replaces all occurrences in a rule. It is used to represent several rules, with a notation. The grammar is still context-free. See \emph{seed rule examples} in Section~\ref{sec:seed}.
		\item $\sigma$ is a set of \emph{terminals}. Terminal is any existing non-null token from any sentence layer. We denote a terminal by \emph{value\{layer\}}. \newline
		For instance, [location]\{named-entity\}, Phil\_Madeira\{instance\}. If the terminal is from the lexical layer, the layer is skipped in the denotation.
		\item $P$ is a set of production rules that represents a relation from $V \rightarrow (V \cup E)^*$. For example, \begin{center}
			\begin{tabular} {c}
				$<$Relation$>$ ::= 	$<$Person$>$ is $<$Life Role$>$ \\
			\end{tabular}
		\end{center}
		%We divide the production rules on lexicon rules, where the right side consists of literals only ( $V \rightarrow E^*$), and pattern rules, where the right side contains at least one non-literal. \todo{skip last sent if not used later}
		\item $S$ is the starting non-terminal symbol. Since non-terminals represent semantic classes, the starting symbol is chosen based on the semantic class of the input examples. If the input examples are sentences, then the appropriate category may be  $\langle \text{Relation} \rangle$. While if the input examples are noun phrases, the starting symbol may be a more specific category, like $\langle \text{Job Title} \rangle$.
		\item $R$ is a set of properties: \emph{positive, neutral, negative, non-inducible}. The property controls the usage of the rule in the parsing and in the rule induction phase. More details are given in the following subsections.
		
	\end{itemize}
	
	\subsection{Parser}
	\label{sec:parser}
	
	For parsing, a recursive descent parser with backtracking was developed. 
	This is a top-down parser, which first looks at the higher level sentence structure and then proceeds down the parse tree to identify low level details of the sentence. 
	The advantage of top-down parsing is the ability to partially parse sentences and to detect unparsable parts of sentences. 
	
\begin{figure}[htbp] 
		\makebox[\textwidth][c]{\includegraphics[width=\textwidth]{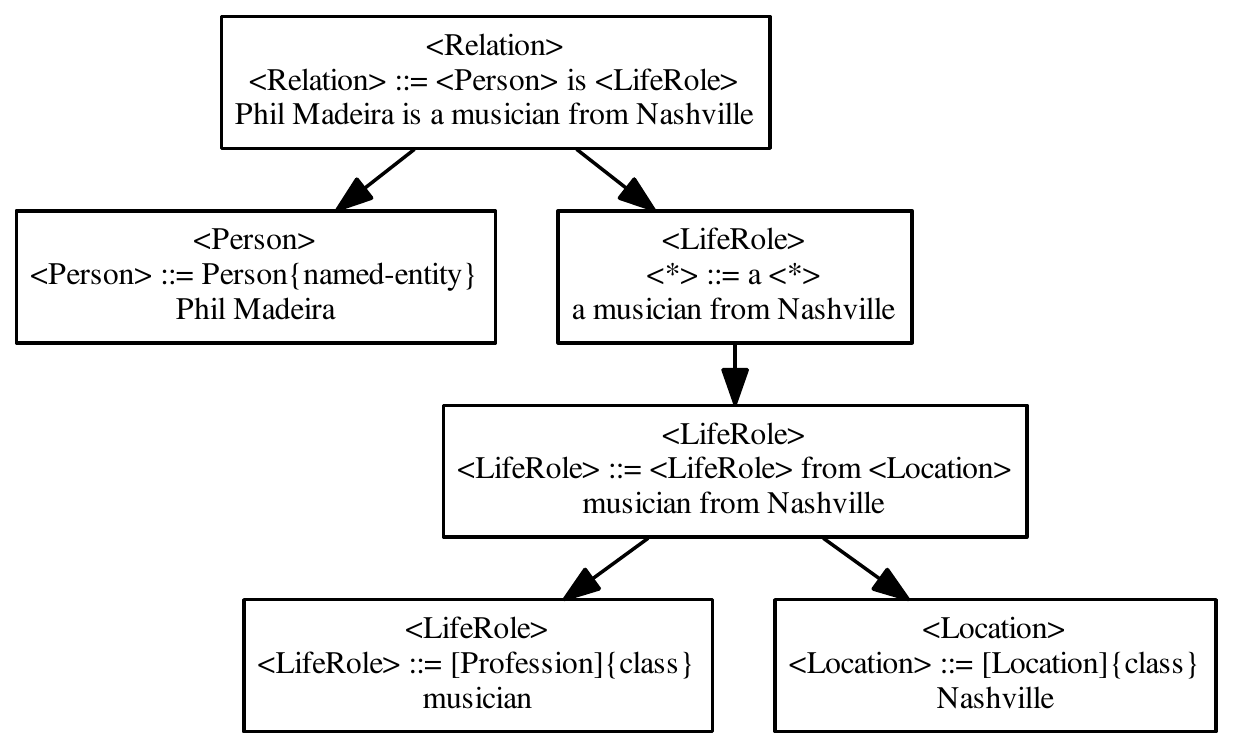}}
		\caption{\small{An example of a semantic tree, which is the result of parsing the input from Table \ref{table:layers}. The tree is fully parsed. Each node has three rows: the class, the rule and the term interpreted in the lexical layer.}}
		\label{semTree} 
	\end{figure}

	The parser takes a layered sentence as an input and returns a semantic tree as an output (see Figure~\ref{semTree}).
	The recursive structure of the program closely follows the structure of the parse tree. 
	The recursive function \emph{Parse} (see Algorithm~\ref{alg:parse}) takes a term and a non-terminal as input and returns a parse node as an output. 
	The parse node contains the class of node (non-terminal), the \emph{rule} that parsed the node, the \emph{term}, and the list of children nodes. 
	In order for the rule to parse the node, the left-hand side must match the input non-terminal and the right-hand side must match the layered input. 
	In the pattern matching function \emph{Match} (line~\ref{alg:match}), the right hand side of a rule is treated like a regular expression; non-terminals present the ($+$) wildcard characters, which match at least one word. 
	The terminals are treated as literal characters, which are matched against the layer that defines them. The result of successfully matched pattern is a list of terms, where each term represents a non-terminal of the pattern. Due to ambiguity of pattern matching there might be several matches. For each of the term -- non-terminal pair in every list the \emph{parse} function is recursively called (line~\ref{alg:rec}).
	
		\begin{algorithm2e}

			\SetKwFunction{GetEligibleRules}{GetEligibleRules}
			\SetKwFunction{CreateNode}{Node}
			\SetKwFunction{Merge}{SelectBestNode}
			\SetKwFunction{Match}{Match}
			\SetKwFunction{Parse}{Parse}
			
			\SetKwData{Grammar}{grammar}
			\SetKwData{Pattern}{pattern}
			\SetKwData{NonTerminals}{non terminals}
			
			\SetKwData{Rules}{rules}
			\SetKwData{ChildTree}{child node}
			
			\SetKwData{Unexpanded}{induction nodes}
			
			\SetKwData{AList}{ambiguous lists}
			\SetKwData{PList}{term list}
			\SetKwData{Child}{child nodes}
			\SetKwData{Final}{final node}
			\SetKwData{Ambiguous}{nodes}
			\SetKwInOut{Input}{input}
			\SetKwInOut{Output}{output}
			
			\SetKwProg{Fn}{Function}
			
		     \Fn{\Parse{Phrase $p$, Non-terminal $n$}} {
				\Output {parse node}
				\BlankLine
				
				\Ambiguous $\leftarrow$ $\{\}$\;
				\ForEach{rule $r$ of \Grammar}
				{
					\If{n = left side of $r$} {
						\Pattern $\leftarrow$ right hand side of $r$\;
						\AList $\leftarrow$ \Match{\Pattern, $p$} 	\label{alg:match} \;
						
						\ForEach {\PList of \AList}
						{
							\Child $\leftarrow$ $\{\}$\;
							\For{$i\leftarrow 0$ \KwTo size of \PList}{
								\ChildTree $\leftarrow$  \Parse{$\PList_i$, $\Pattern.\NonTerminals_i$} 	\label{alg:rec}\; 
								add \ChildTree to \Child \;
							}
							add \CreateNode{$type$, $p$ , $r$, \Child} to \Ambiguous\;
						}		
					}
					
				}
				
				\BlankLine
				\If{\Ambiguous is empty} {
					\Final $\leftarrow$ \CreateNode{$type$, $p$ , null, $\{\}$}\;	
				}
				\Else {
					\Final $\leftarrow$ $\argmax_{n \in nodes} r(n)$  \label{alg:rel}\;
				}
				\If{\Final is not fully parsed} {
					add \Final to \Unexpanded \label{alg:ind_nodes}\; 
				}
				\KwRet {\Final}
				\BlankLine	
				\BlankLine
			}
			\caption{Pseudocode of the main function \emph{parse} of the top-down parser.}
			\label{alg:parse}
			
		\end{algorithm2e}  
	
	Since the grammar is ambiguous, a term can be parsed in multiple ways. 
	There are two types of ambiguity. Two or more rules can expand the same term and one rule can expand the term in more than one way. For each ambiguity one node is created, and the best node according to \emph{reliability measure} is selected to be the result (line~\ref{alg:rel}). The reliability measure $r(n)$ is
	\begin{equation}
	r(n)= 
	\begin{cases}
	1,              & \text{if node is fully parsed} \\
	\beta \cdot (1 -tp(n)) + (1 - \beta)\frac{\displaystyle\sum\limits_{c \in C(n)} |c|\cdot r(c)}{\displaystyle\sum\limits_{c \in C(n)} |c|} ,& \text{if node is partially parsed} \\
	0,				 & \text{if node is null} \\
	\end{cases}
	\label{eq:reliability}
	\end{equation}
	where $tp(n)$ is the trigger probability of the rule that parsed the node $n$, $\beta$ is a predefined weight, $C(n)$ is the set of children of $n$, and $|c|$ is the length of the term of node $c$.
	The trigger probability of the rule is the probability that a the right-hand side of the rule pattern matches a random term in the dataset and it is estimated after the rule is induced. The range of the measure is between $0$ and $1$. The measure was defined in such a way that the more text the node parses, the higher is the reliability (the second summand in the middle row of Eq.~\ref{eq:reliability}). On the other hand, nodes with rules that are more frequently matched have lower reliability; this penalizes rules that are very loosely defined (the first summand in the middle row of Eq.~\ref{eq:reliability}). The $\beta$ parameter was set to 0.05, using grid search, with average F1 score from relation extraction experiment from Section~\ref{sec:experiment_relation} as a performance measure. 
	
	If none of the rules match the term, a \emph{null node} is created and added to the list of nodes, which will be later used for grammar induction (line~\ref{alg:ind_nodes}). 
	Note that even if a null node is discarded, because it is not the most reliable, it will still be used in the grammar induction step.
	A node is \emph{fully parsed} if the node itself and all of its descendants are parsed. If a node is parsed and if at least one of its descendants is not parsed, then the node is \emph{partially parsed}. 
	All nodes that are not fully parsed are added to the list for induction.
	
	Since the ambiguity of the grammar may make parsing computationally infeasible, several optimization techniques are used. Memoization \cite{Norvig:1991:TAM:971738.971743} is used to reduce the complexity from exponential time to $\mathcal{O}(n^3)$ \cite{Frost:2006:NTP:1149982.1149988}, where $n$ is the length of the sentence. 
	The parser does not support $\epsilon$ productions mainly because the grammar induction will not produce them. 
	The patterns that do not contain terminals are the most ambiguous. 
	At most two non-terminals are allowed, and the maximal length of the term that corresponds to the first non-terminal is three tokens. We argue that this is not a huge limitation, since the way human languages are structured, usually two longer terms are connected with a word, like comma or a verb. Furthermore, the way how our induction works, these connectors do not get generalized and become a terminal in the rule.
	There was an attempt to introduce rules with negative property. Whenever such rule fully parses a node, that indicates that the current parsing path is incorrect. This allows the parser to backtrack sooner and also prevents adding null sister nodes (null sister nodes are in this case usually wrong) to the rule induction. However, it turned out that negative rules actually slow down the parsing, since the grammar gets bigger. It is better to mark these rules as neutral, therefore they are not added to the grammar.

	\subsection{Rule induction}
	\label{sec:rule_induction}
	
	The goal of the rule induction step is to convert the null nodes from the parsing step into rules. Out of these rules, the \emph{most frequent} one is promoted.
	The term from the null node is \emph{generalized} to form the right side of the rule. The class non-terminal of the null node will present the left side of the rule. 
	Recently induced rule will parse all the nodes, from which it was induced, in the following iterations. 
	Additionally, some rules may parse the children of those nodes.  
	
	\paragraph{Generalization}  \hspace{0pt} \\
	Generalization is done in two steps. First, terms are generalized on the layer level. The output of this process is a sequence of tokens, which might be from different layers. For each position in the term a single layer is selected, according to predefined layer order. In the beginning, term is generalized with the first layer. All the non-null tokens from this layer are taken to be part of the generalized term. All the positions of the term that have not been generalized are attempted to be generalized with the next layer, etc. The last layer is without null-tokens, therefore each position of the term is assigned a layer. Usually, this is the lexical layer. 
	For example, top part of Table~\ref{tab:generalization} shows generalization of term from Table~\ref{table:layers}. 
	The layer list is constructed manually. Good layers for generalization are typically those that express semantic classes of individual terms. Preferably, these types are not too general (loss of information) and not too specific (larger grammar). 
	
	In the next step of generalization, tokens are further generalized using a greedy bottom-up parser using the rules from the grammar. The right sides of all the rules are matched against the input token term. If there is a match, the matched sub-term is \emph{replaced} with the left side of the rule. Actually, in each iteration all the disjunct matches are replaced. To get only the disjunct matches, overlapping matches are discarded greedily, where longer matches have the priority. This process is repeated until no more rules match the term. An example is presented in the lower part of Table~\ref{tab:generalization}.
	
	The bottom-up parsing algorithm needs to be fast because the number of unexpanded nodes can be very high due to ambiguities in the top-down parsing. Consequently, the algorithm is greedy, instead of exhaustive, and yields only one result. Aho-Corasick string matching algorithm \cite{Aho:1975:ESM:360825.360855} is selected for matching for its ability to match all the rules simultaneously.
	Like the top-down parser, this parser generates partial parses because the bottom-up parser will never fully parse -- the output is the same as the non-terminal type in the unexpanded node. This would generate a cyclical rule, i.e. \emph{$<$Class$>$ :== $<$Class$>$}. However, this never happens because the top-down parser would already expand the null node. 
	
	\begin{table}
		\centering

		\begin{tabular} {|c|c|c|c|c|c|}
			\hline
			\multicolumn{6}{|c|}{\textbf{Layered generalization}}\\
			\hline
			Person &is  &a &Profession &from & Location \\
			class & small-caps & small-caps & class & small-caps & class \\
			\hline
			\multicolumn{6}{}{}\\
			\hline
			\multicolumn{6}{|c|}{\textbf{Bottom-up parsing}}\\
			\hline
			Person &is  &a &\multicolumn{3}{|c|}{$<$Life Role$>$} \\
			class & small-caps & small-caps &\multicolumn{3}{|c|}{non-terminal} \\
			\hline
		\end{tabular}
		\caption{\small{Two step generalization of term from Table~\ref{table:layers} (Phil Madiera is a musician from Nashville). The layer list was constructed in a reverse order from Table~\ref{table:layers}. In this example, the rule \emph{$<$Life Role$>$ ::= Profession from Location} is the only rule that was used in the bottom-up parsing. }}
		\label{tab:generalization}
	\end{table}
	
	\paragraph{Property assignment} \hspace{0pt} \\
	The last step of the iteration is assigning the property to the newly induced rule. 
	Property controls the role of the rule in the parsing and induction. The default property is \emph{positive}, which defines the default behavior of the rule in all procedures. Rules with \emph{neutral} property are not used in any procedure. They also cannot be re-induced. Some rules are good for parsing, but may introduce errors in the induction. These rules should be given \emph{non-inducible} property. For instance, rule \emph{$<$Date$>$ :== $<$Number$>$} is a candidate for the non-inducible property, since years are represented by a single number. On the contrary, not every number is a date.
	
	In our experiments, the assignment was done manually.
	The human user sees the induced rule and few examples of the null nodes, from which it was induced. 
	This should provide enough information for the user to decide in a few seconds, which property to assign. After the stopping criteria is met, the iterative procedure can continue automatically by assigning positive property to each rule.
	Initial experimenting showed that just a single mistake in the assignment can cause a huge drift, making all further rules wrong.

	\subsection{Seed rules}
	\label{sec:seed}
	Before the start, a list of seed rules may be needed in order for grammar induction to be successful. 
	Since this step is done manually, it is reasonable to have a list of seed rules short and efficient. 
	Seed rules can be divided in three groups: domain independent linguistic rules, class rules, top-level domain rules. 
	\emph{Domain independent linguistic rules}, such as 
	\begin{center}
		\begin{tabular} {l l}
			$<$\_$>$ ::= a $<$\_$>$ & $<$Relation $>$ ::=  $<$Relation$>$ . \\
			$<$\_$>$ ::= an $<$\_$>$ &  $<$\_$>$ ::=$<$\_$>$ and $<$\_$>$ \\
			$<$\_$>$ ::= the $<$\_$>$ &  $<$\_$>$ ::= $<$\_$>$ , $<$\_$>$ and $<$\_$>$,\\ 
		\end{tabular}
	\end{center}
	parse the top and mid-level nodes. They can be applied on many different datasets.
	\emph{Class rules} connect class tokens, like named-entity tokens with non-terminals. For example,
	\begin{center}
		\begin{tabular} {l l}
			$<$Location$>$ ::= [location]\{named-entity\}  & $<$Date$>$ ::= [date]\{named-entity\} \\
			$<$Location$>$ ::= [Location]\{class\}  & $<$Film$>$ ::= [Film]\{class\} \\
		\end{tabular}
	\end{center}
	They parse the leaf nodes of the trees. 
	On the other hand, \emph{top-level domain rules}, define the basic structure of the sentence. For example,
	\begin{center}
		\begin{tabular} {c}
			$<$Relation$>$ ::= 	$<$Person$>$ is $<$Life Role$>$ \\
		\end{tabular}
	\end{center}
	As the name suggests, they parse nodes close to the root. Altogether, these rule groups parse on all levels of the tree, and may already be enough to parse the most basic sentences, but more importantly, they provide the basis for learning to parse more complex sentences. 
	
	The decision on which and how many seed rules should be defined relies on human judgment whether the current set of seed rules is powerful enough to ignite the bootstrapping procedure. This judgment may be supported by running one iteration and inspecting the top induced rules.    
	
	\section{Ontology induction}
	\label{sec:onto_induction}
	
	This section describes how to utilize the grammar and manipulate semantic trees to discover ontology components in the textual data.
	
	\subsection{Ontology induction from grammar}
	\label{sec:from_grammar}
	We propose a procedure for mapping grammar components to ontology components. 
	In particular, classes, instances and taxonomic relations are extracted. 
	
	First, we distinguish between instances and classes in the grammar. 
	Classes are represented by all non-terminals and terminals that come from a layer populated with classes, for example, \emph{named-entity} layer and \emph{class} layer from Table~\ref{table:layers}. 
	Instances might already exist in the \emph{instance} layer, or they are created from rules, whose right hand side contains only tokens from the \emph{lexical} layer. These tokens represent the label of the new instance. For instance rule \emph{$<$Profession$>$ ::= software engineer} is a candidate for instance extraction. 
	
	Furthermore, we distinguish between class and instance rules. 
	\emph{Class rules} have a single symbol representing a class on the right-hand side. 
	Class rules map to \emph{subClassOf} relations in the ontology. If the rule is positive, then the class on the right side is the subclass of the class on the left side. For instance, rule \emph{$<$Organization$>$ ::= $<$Company$>$} yields relation \emph{(subClassOf Company Organization)}.
	
	On the other hand, \emph{instance rules} have one or more symbols representing an instance on the right side, and define the \emph{isa} relation. If the rule is positive, then the instance on the right side is a member of a class on the left side. For instance, rule \emph{$<$Profession$>$ ::= software engineer} yields relation \emph{(isa SoftwareEngineer Profession)}.
	If class or instance rule is neutral then the relation can be treated as false.
	Note that many other relations may be inferred by combing newly induced relations and relations from the seed ontology. For instance, induced relation (\emph{subClassOf new-class seed-class}) and seed relation (\emph{isa seed-class seed-instance}) are used to infer a new relation (\emph{isa new-class seed-instance}).
	
	In this section, we described how to discover relations on the taxonomic level. In the next section, we describe how to discover relations between instances.
	
	\subsection{Relation extraction from semantic trees}
	\label{sec:tree_rule_learning}
	
	We propose a method for learning relations from semantic trees, which tries to solve the same problem as the classical relation extraction methods. 
	Given a dataset of positive relation examples that represent one relation type, e.g. \emph{birthPlace}, the goal is to discover new unseen relations.
	
	The method is based on the assumption that a relation between entities is expressed in the shortest path between them in the semantic tree \cite{Bunescu:2005:SPD:1220575.1220666}.
	The input for training are sentences in layered representation, corresponding parse trees, and relation examples. Given a relation from the training set, we first try to identify the sentence containing each entity of the relation. The relation can have one, two, or even more entities. Each entity is matched to the layer that corresponds to the entity type. For example, strings are matched to the lexical layer;  ontology entities are matched to the layer containing such entities. The result of a successfully matched entity is a sub-term of the sentence. In the next step, the corresponding semantic tree is searched for a node that contains the sub-term.
	
\begin{figure}
		\centering
		\begin{subfigure}[b]{.33\textwidth}
			\centering
			\includegraphics[width=.9\linewidth]{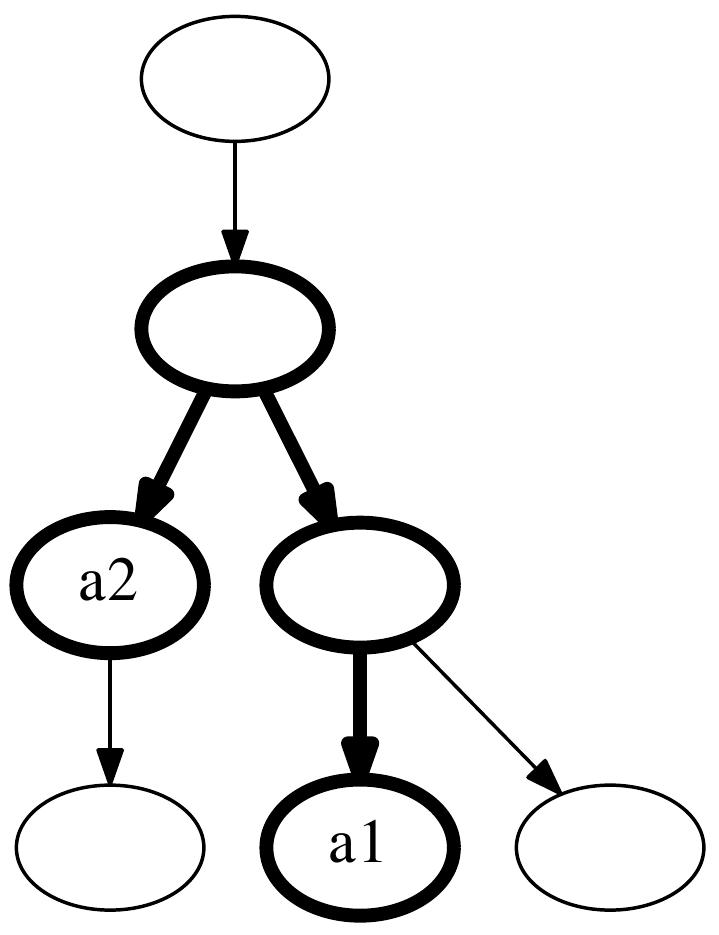}
			\caption{Semantic tree for training, with $a1$ and $a2$ as entity nodes.}
			\label{fig:sub1}
		\end{subfigure}
		\begin{subfigure}[b]{.27\textwidth}
			\centering
			\includegraphics[width=.9\linewidth]{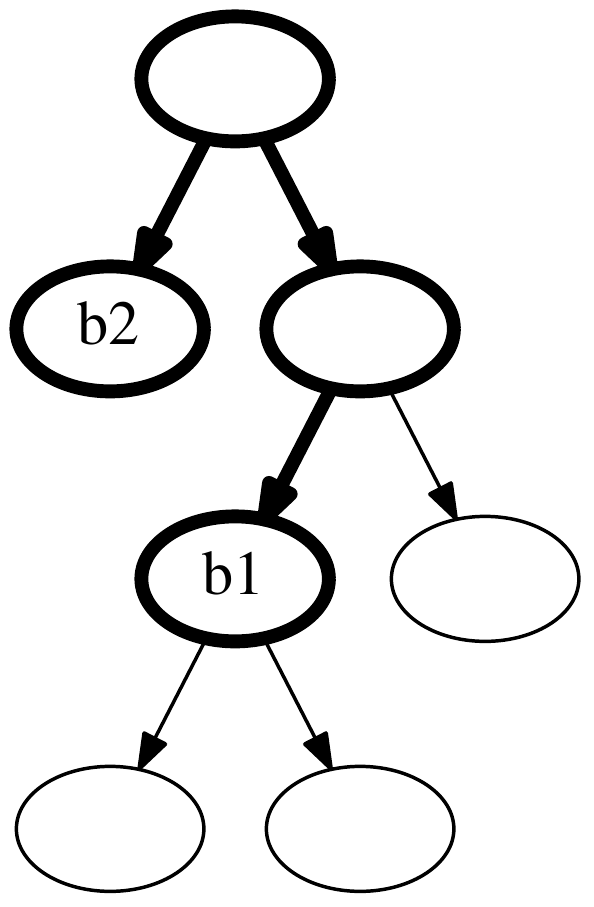}
			\caption{Semantic tree for training, with $b1$ and $b2$ as entity nodes.}
			\label{fig:sub2}
		\end{subfigure}
		\begin{subfigure}[b]{.3\textwidth}
			\centering
			\includegraphics[width=.8\linewidth]{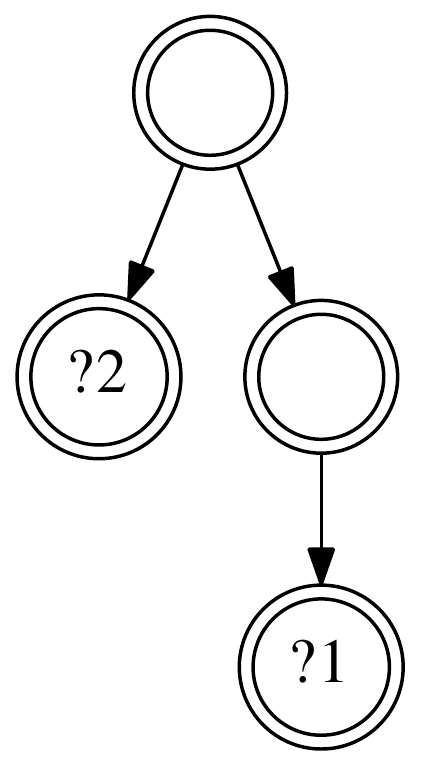}
			\caption{Variable sub-tree, $?1$ and $?2$ present the positions in the relation.}
			\label{fig:sub3}
		\end{subfigure}
		\caption{Given the relation $(r~a1~a2)$ and semantic tree on \ref{fig:sub1}, and relation $(r~b1~b2)$ and semantic tree on \ref{fig:sub2}, the same variable sub-tree is extracted (\ref{fig:sub3}).}
		\label{fig:var_trees}
	\end{figure}

	At this point, each entity has a corresponding \emph{entity node}.
	Otherwise, the relation is discarded from the learning process. 
	Given the entity nodes, a minimum spanning tree containing all off them is extracted. 
	If there is only one entity node, then the resulting subtree is the path between this node and the root node. 
	The extracted sub-tree is converted to a \emph{variable tree}, so that different semantic trees can have the same variable sub-trees, for example see Figure~\ref{fig:var_trees}. 
	The semantic nodes of the sub-tree are converted into variable nodes, by retaining the class and the rule of the node, as well as the places of the children in the original tree. For entity nodes also the position in the relation is memorized. 
	Variable tree extracted from a relation is a positive example in the training process. 
	For negative examples all other sub-trees that do not present any relations are converted to variable trees.
	Each variable node represents one feature. 
	Therefore, a classification algorithm, such as logistic regression can be used for training.
	
	When predicting, all possible sub-trees of the semantic tree are predicted\footnote{The number of leaf nodes of these sub-trees must match the number of arguments of the relation. Also, if the relation has two or mode arguments, it is predicted several times, each time with the different position numbers in the entity nodes.}. If a sub-tree is predicted as positive, then the terms in the leaf nodes represent the arguments of the relation. 
	
	\section{Experiments}
	\label{sec:experiments}
	In this section, we present experiments evaluating the proposed approach. We have conducted experimentation on Wikipedia--DBpedia dataset (Section~\ref{sec:datasets}). First, we have induced a grammar on the Wikipedia dataset (Section~\ref{sec:induction_experiments}) to present its characteristics, and the scalability of the approach. In the next experiment, we present a method for discovering less prominent instances (Section~\ref{sec:inst_learning}). The last experiment demonstrates one application of semantic parsing -- the supervised learning of DBpedia relations(Section~\ref{sec:experiment_relation}).   
	
	\subsection {Datasets}
	\label{sec:datasets}
	
	The datasets for experiments were constructed from English Wikipedia and knowledge bases DBpedia \cite{dbpedia-swj} and Freebase \cite{Bollacker:2008:FCC:1376616.1376746}. DBpedia provides structured information about Wikipedia articles that was scraped out of their infoboxes.
	First sentences of Wikipedia pages describing people were taken as the \emph{textual dataset}, while DBpedia relations expressing facts about the same people were taken as the \emph{dataset for supervised relation learning}. Note that each DBpedia instance has a Wikipedia page. A set of person instances was identified by querying DBpedia for instances that have a person class. 
	For the textual dataset, Wikipedia pages representing these entities were parsed by the in-house Wikipedia markup parser\footnote{The markup parsing of Wikipedia markup is non-trivial. To extract only the plain text many elements, like images and tables, have been discarded, as well as some elements that appear in the middle of sentence, like pronunciation and citation} to convert the markup into plain text. 
	Furthermore, the links to other Wikipedia pages were retained. Here is an example of a sentence in plain text:
	
	\begin{quote}
		\emph{\blockquote{Victor Francis Hess (24 June 1883 – 17 December 1964) was an Austrian-American physicist, and Nobel laureate in physics, who discovered cosmic rays.}}
	\end{quote}
	
	Using the Standford OpenNLP \cite{manning-EtAl:2014:P14-5} on plain texts we obtained sentence and token splits, and named-entity annotation. Notice, that only the first sentence of each page was retained and converted to the proposed layered representation (see Section~\ref{sec:textdata}). 
	The layered representation contains five layers: \emph{lexical} (plain text), \emph{named-entity} (named entity recognizer), \emph{wiki-link} (Wikipedia page in link -- DBpedia instance\footnote{Each Wikipedia page has a corresponding Dbpedia instance.}), \emph{dbpedia-class} (class of Wikipedia page in Dbpedia) and \emph{freebase-class} (class of Wikipedia page in Freebase).
	Freebase also contains its own classes of Wikipedia pages. 
	For the last two layers, there might be several classes per Wikipedia page. Only one was selected using a short priority list of classes. If none of the categories is on the list, then the category is chosen at random. After comparing the \emph{dbpedia-class} and \emph{freebase-class} layers, only \emph{freebase-class} was utilized in the experiments because more \emph{wiki-link} tokens has a class in \emph{freebase-class} layer than in \emph{dbpedia-class} layer.
	
	There are almost 1.1 million sentences in the collection. The average length of a sentence is 18.3 words, while the median length is 13.8 words. There are 2.3 links per sentence.
	
	The dataset for supervised relation learning contains all relations where a person instance appears as a subject in DBpedia relation. For example, 
	\begin{quote}
		dbpedia:Victor\_Francis\_Hess ~~~ dbpedia-owl:birthDate ~~~ 1883-06-24
	\end{quote}
	There are 119 different relation types (unique predicates), having from just a few relations to a few million relations. Since DBpedia and Freebase are available in RDF format, we used the RDF store for querying and for storage of existing and new relations.    
	
	\subsection {Grammar Induction Experiments}
	\label{sec:induction_experiments}
	
	The grammar was induced on 10.000 random sentences taken from the dataset described in Section~\ref{sec:datasets}. First, a list of 45 seed nodes was constructed. There were 22 domain independent linguistic rules, 17 category rules and 6 top-level rules.
	The property assignment was done by the authors. 
	In every iteration, the best rule is shown together with the number of nodes it was induced from, and ten of those nodes together with the sentences they appear in. The goal was set to stop the iterative process after two hours. We believe this is the right amount of time to still expect quality feedback from a human user. 
	
	There were 689 new rules created. A sample of them is presented in Table~\ref{tab:rules}. Table~\ref{table:gstats} presents the distributions of properties. Around $36 \%$ of rules were used for parsing (non neutral rules). Together with the seed rules there are  297 rules used for parsing. Different properties are very evenly dispersed across the iterations. Using the procedure for conversion of grammar rules into taxonomy presented in Section~\ref{sec:onto_induction}, \textbf{33} classes and subClassOf relations, and \textbf{95} instances and isa relations were generated. 
	
\begin{figure*}
		\centering
		\begin{subfigure}[b]{0.475\textwidth}
			\centering
			\includegraphics[width=\textwidth]{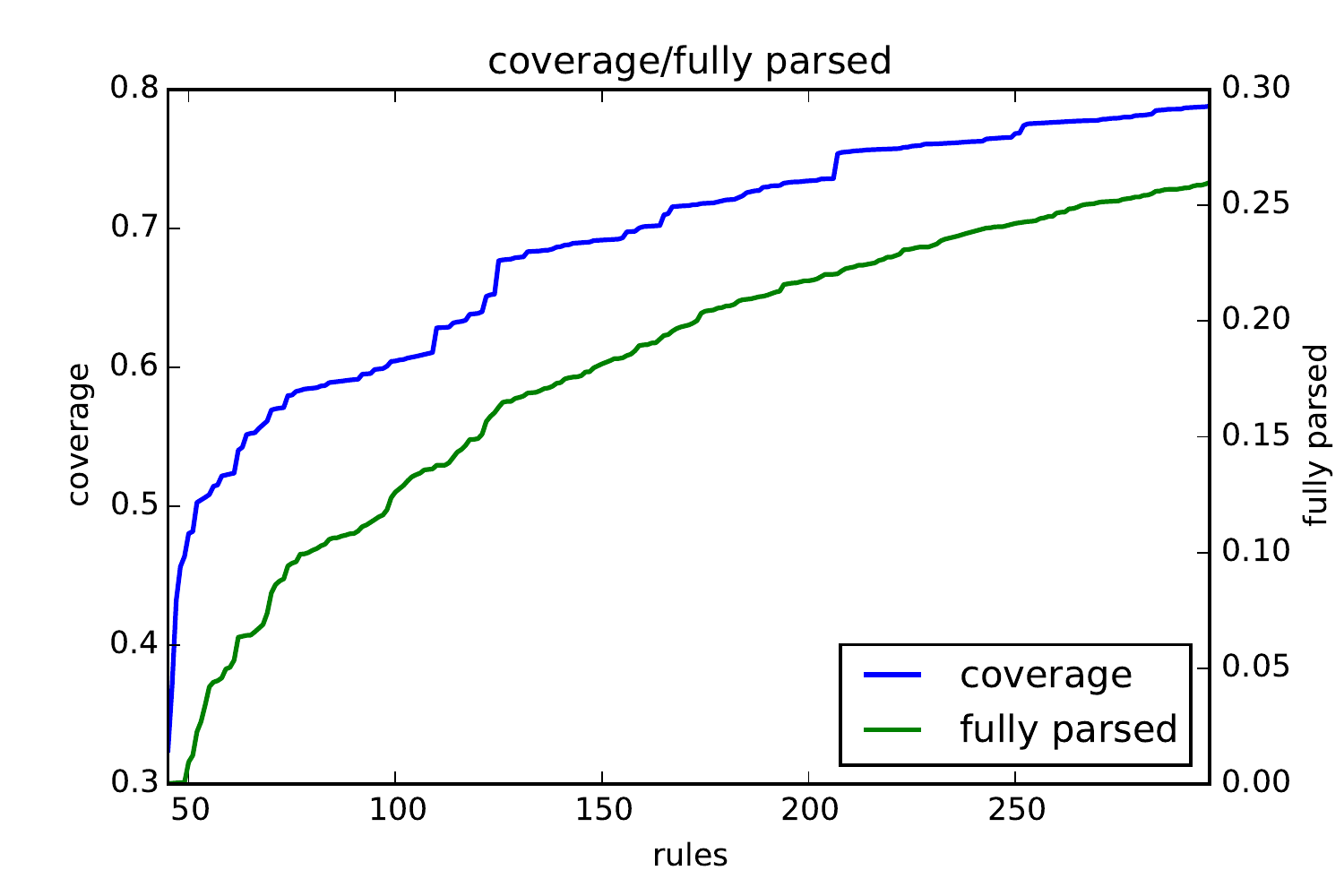}
			
			\caption{Coverage/fully parsed}
			\label{fig:coverage}
		\end{subfigure}
		\hfill
		\begin{subfigure}[b]{0.475\textwidth}  
			\centering 
			\includegraphics[width=\textwidth]{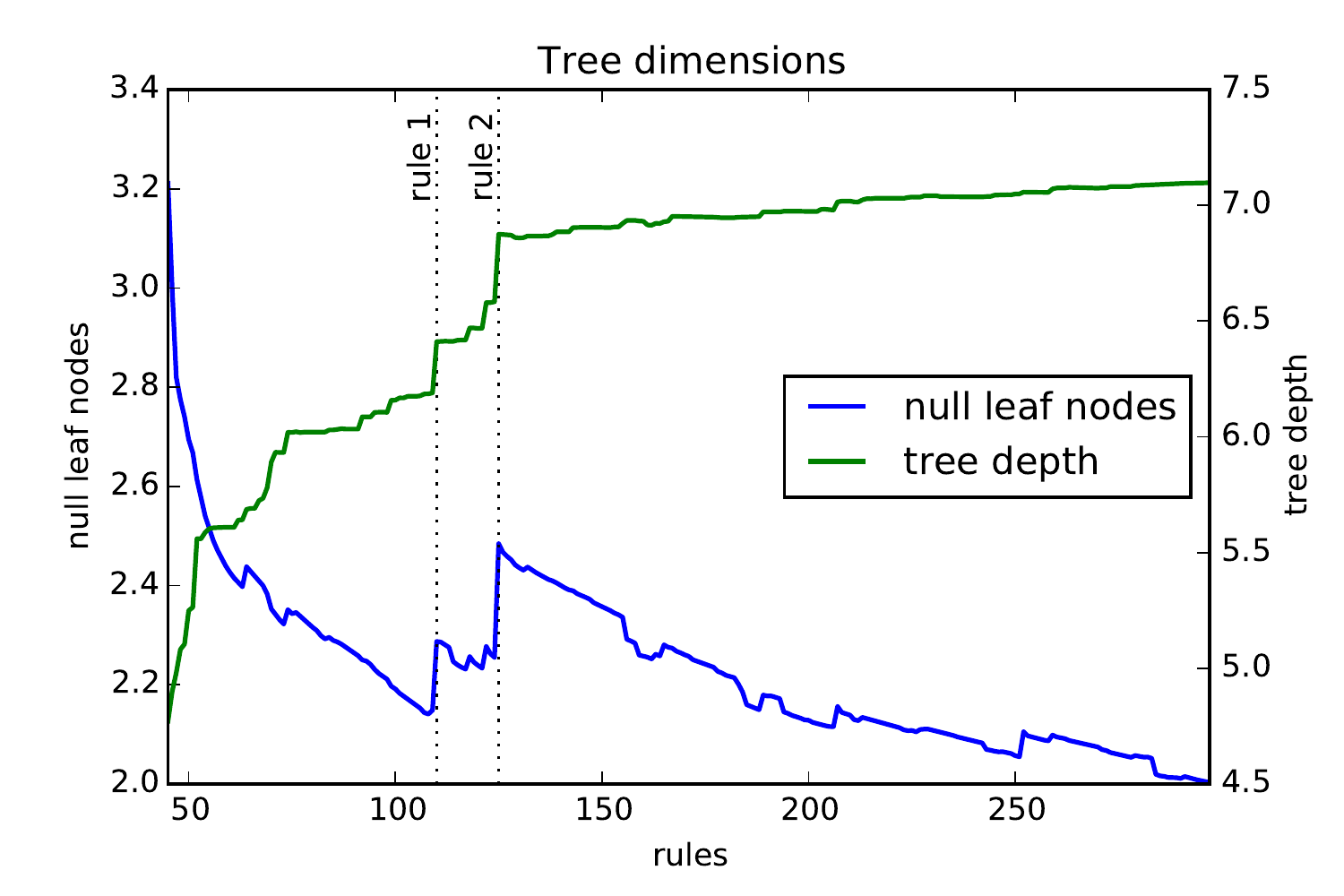}
			\caption{Tree dimensions}
			\label{fig:nodes}
		\end{subfigure}
		\vskip\baselineskip
		\begin{subfigure}[b]{0.475\textwidth}   
			\centering 
			\includegraphics[width=\textwidth]{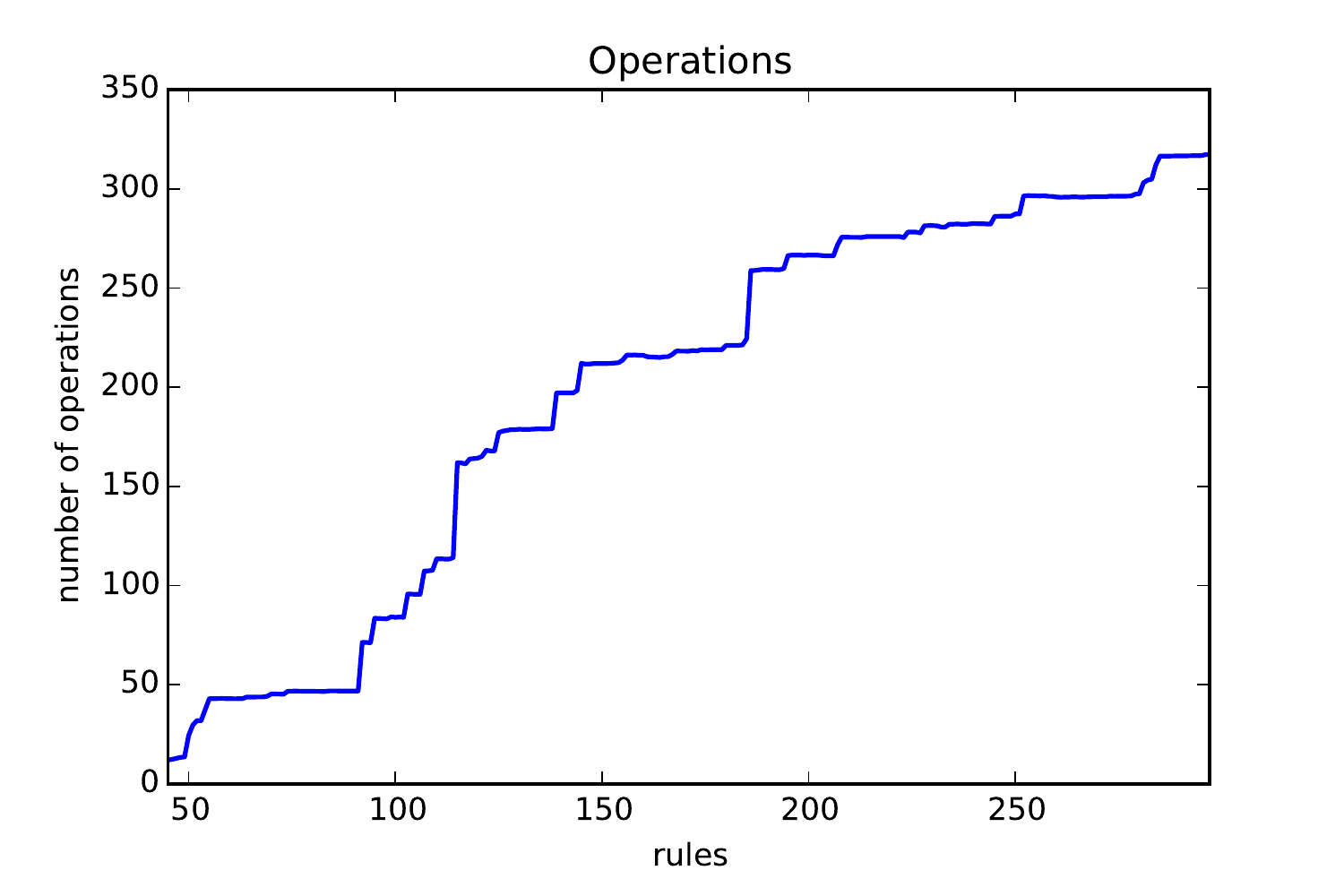}
			\caption{Operations}
			\label{fig:operations}
		\end{subfigure}
		\quad
		\begin{subfigure}[b]{0.475\textwidth}   
			\centering 
			\includegraphics[width=\textwidth]{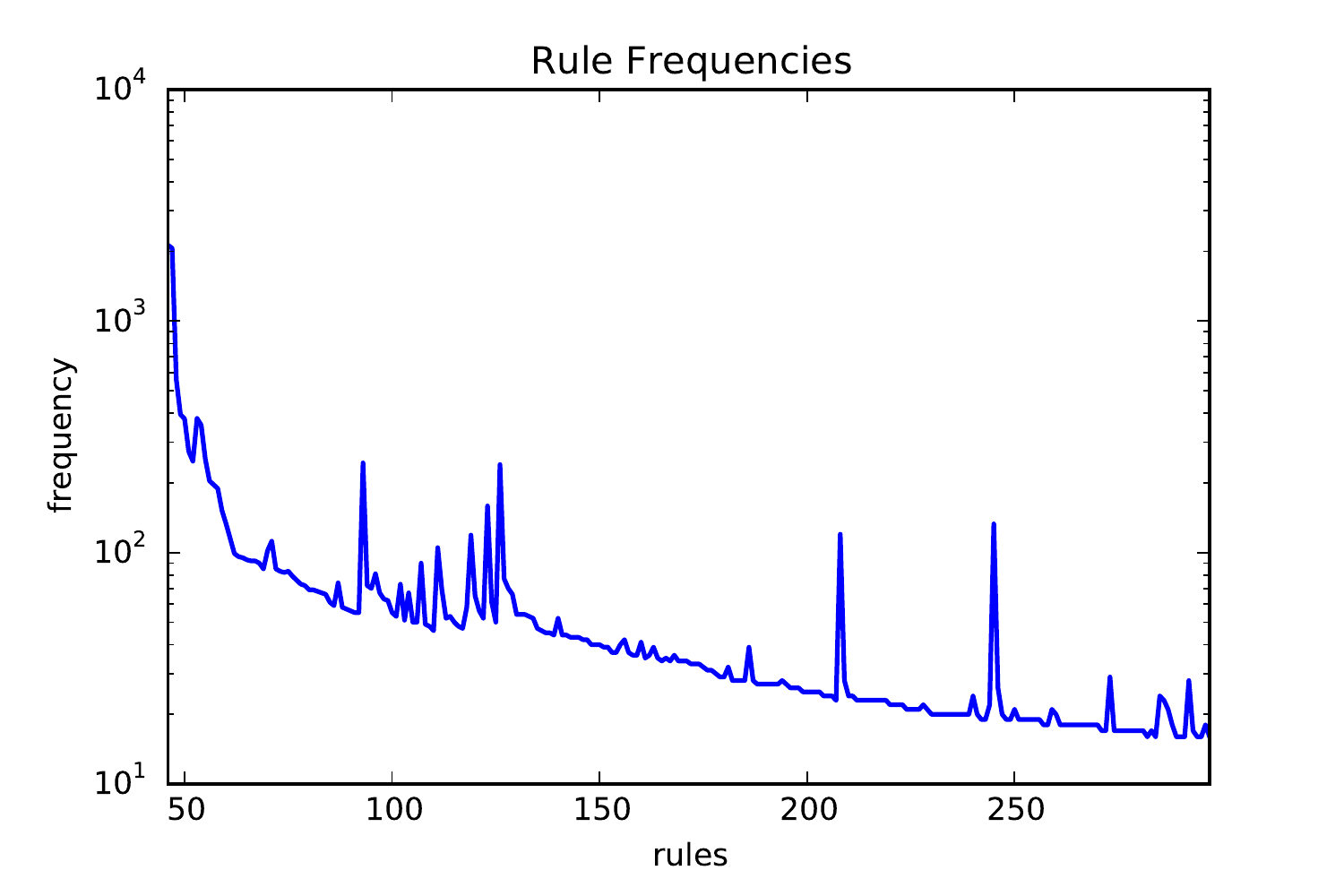}
			\caption{Rule frequencies}
			\label{fig:score}
		\end{subfigure}
		\caption{Statistics of the iterative process} 
		\label{fig:iterations}
	\end{figure*}

	The grammar was also tested by parsing a sample of 100.000 test sentences. A few statistic are presented in Table~\ref{table:gstats}. More than a quarter of sentences were fully parsed, meaning that they do not have any null leaf nodes. Coverage represents the fraction of words in a sentence that were parsed (words that are not in null-nodes).
	%Since the average depth of a tree is larger than the average number of leaf nodes shows that many rules have only one right-side non-terminal.
	The number of operations shows how many times was the \emph{Parse} function called during the parsing of a sentences. It is highly correlated with the time spend for parsing a sentence, which is on average 0.16ms. This measurement was done on a single CPU core. Consequently, it is feasible to parse a collection of a million sentences, like our dataset. The same statistics were also calculated on the training set, the numbers are very similar to the test set. The fully parsed \% and coverage are even slightly lower than on the test set. Some of the statistics were calculated after each iteration, but only when a non neutral rule was created. The graphs in Figure~\ref{fig:iterations} show how have the statistics changed over the course of the grammar induction.  Graph~\ref{fig:coverage} shows that coverage and the fraction of fully parsed sentences are correlated and they grow very rapidly at the beginning, then the growth starts to slow down, which indicates that there is a long tail of unparsed nodes/sentences. In the following section, we present a concept learning method, which deals with the long tail. Furthermore, the number of operations per sentence also slows down (see Graph~\ref{fig:operations}) with the number of rules, which gives a positive sign of retaining computational feasibility with the growth of the grammar. Graph~\ref{fig:nodes} somewhat elaborates the dynamics of the grammar induction. In the earlier phase of induction many rules that define the upper structure of the tree are induced. These rules can rapidly increase the depth and number of null nodes, like \emph{rule 1} and \emph{rule 2}
	\footnote{rule1 -- \emph{$<$LifeRole$>$ ::=  $<$LifeRole$>$ in $<$Location$>$}, \\ 
		~~~rule2 -- 	\emph{ $<$LifeRole$>$ ::=  $<$LifeRole$>$ of $<$Organization$>$ }}. They also explain the spikes on Graph~\ref{fig:score}. Their addition to the grammar causes some rules to emerge on the top of the list with a significantly higher frequency. After these rules are induced the frequency gets back to the previous values and slowly decreases over the long run.

	\begin{table}
		\centering

		\begin{tabular}{lll}
			
			\hline
			\textbf{number} & \textbf{rule} & \textbf{property} \\
			
		\hline
			1	& $<$PersonAttr$>$ ::= born $<$Date$>$ &	none \\
			101 & $<$LifeRole$>$ ::= born in $<$Location$>$&	none\\
			201	&  $<$Location$>$ ::= $<$Location$>$ from $<$Date$>$&	neutral\\
			301	&  $<$Person$>$ ::= $<$Location$>$ from $<$Date$>$&	neutral\\
			401	&  $<$OrderOfChivalry$>$ ::= ( $<$PersonAttr$>$ )	&neutral\\
			501	&  $<$LifeRole$>$ ::= who $<$Action$>$ in $<$Event$>$	&none\\
			601	&  $<$Date$>$ ::= the university&	neutral\\
		\hline
		\end{tabular}
		
		\caption{A sample of induced rules.}
		\label{tab:rules}
	\end{table}
	
	\begin{table}
		\centering
			
		\begin{tabular}{ll}
			
			\hline
			\multicolumn{2}{l}{\textbf{Grammar}} \\
			\hline
			positive rules & 231 \\
			non-inducible rules & 21 \\
			neutral rules & 437 \\
			\hline\noalign{\smallskip}
			\hline\noalign{\smallskip}
			\multicolumn{2}{l}{\textbf{Parsing}} \\
		    \hline
			fully parsed sentences & $25.63\%$ \\
			avg. coverage & $78.52\%$ \\
			avg. tree depth & 6.96 \\
			avg. number of leaf nodes & 6.69 \\
			avg. number of null leaf nodes  & 1.98 \\
			avg. number of operations & 320.3 \\
			avg parsing time & 0.16 ms\\
			\hline
			
		\end{tabular}
		
		\caption{Statistics of test set.}
		\label{table:gstats}
	\end{table}
	
	\subsection{Instance extraction}
	\label{sec:inst_learning}
	In this section, we present an experiment with a method for discovering new instances, which appear in the long tail of null nodes.
	Note that the majority of the instances were already placed in the ontology by the method in Section~\ref{sec:from_grammar}. Here, less prominent instances are extracted to increase the coverage of semantic parsing.
	The term and the class of the null node will form an \emph{isa} relation. The class of the node represents the class of the relation. The terms are converted to instances. They are first generalized on the layer level (see Section~\ref{sec:textdata}). The goal is to exclude non-atomic terms, which do not represent instances. Therefore, only terms consisting of one \emph{wiki-link} token or exclusively of lexical tokens are retained. The relations were sorted according to their frequency. We observe that accuracy of the relations drops with the frequency. Therefore, relations that occurred less than three times were excluded. The number and accuracy for six classes is reported in Table~\ref{tab:concepts}. Other classes were less accurate. For each class, the accuracy was manually evaluated on a random sample of 100 instance relations. Taking into account the estimated accuracy, there were more than 13.000 correct \emph{isa} relations.
	
	\begin{table}
		\centering

		\begin{tabular}{lll}
			
			\hline\noalign{\smallskip}
			Class & Relations & Accuracy \\
			
			\noalign{\smallskip}\hline\noalign{\smallskip}
			Life role & 15356 &	60 \\
			Person & 5427 & 49 \\
			Order of chivalry & 1319 & 32 \\
			Date & 1310 & 46 \\
			Action & 967 & 78  \\
			Field of study & 25 & 80 \\		
			\hline\noalign{\smallskip}
		\end{tabular}
		
		\caption{Instance relations per category.}
		\label{tab:concepts}
	\end{table}

	\subsection {Relation extraction}
	\label{sec:experiment_relation}
	In this section, we present an experiment of the relation extraction methods presented in Section~\ref{sec:tree_rule_learning}. 
	The input for the supervision is the DBpedia relation dataset from Section~\ref{sec:datasets}. The subject (first argument) of every relation is a person DBpedia instance -- person Wikipedia page. 
	In the beginning, the first sentence of that Wikipedia page has been identified in the textual dataset. 
	If the object (last argument) of this relation matches a sub-term of this sentence, then the relation is eligible for experiments. 
	We distinguish three types of values in objects. DBpedia resources are matched with \emph{wiki-link} layer. Dates get converted to the format that is used in English Wikipedia. They are matched against the \emph{lexical} layer, and so are the string objects.
	
	Only relation types that have 200 or more eligible relations have been retained. This is 74 out of 119 relations. The macro average number of eligible relations per relation type is 17.7\%. While the micro average is 23.8\%, meaning that roughly a quarter of all DBpedia person relations are expressed in the first sentence of their Wikipedia page. For the rest of this section, all stated averages are micro-averages.
	
	The prediction problem is designed in the following way. Given the predicate (relation type) and the first argument of the relation (person), the model predicts the second argument of the relation (object). Because not all relations are functional, like for instance \emph{child} relation, there can be several values per predicate--person pair; on average there are 1.1. 
	Since only one argument of the relation is predicted, the variable trees presented in Section~\ref{sec:tree_rule_learning}, will be paths from the root to a single node.
	Analysis of variable tree extraction shows that on average 60.8\% of eligible relations were successfully converted to variable trees (the object term exactly matches the term in the node). Others were not converted because 8.2\% of the terms were split between nodes and 30.9\% terms are sub-terms in nodes instead of complete terms.
	Measuring the diversity of variable trees shows that a distinct variable tree appeared 2.7 times on average.
	
\begin{center}
		\begin{figure}[ht]
			\centering
			\includegraphics[width=1.0\textwidth]{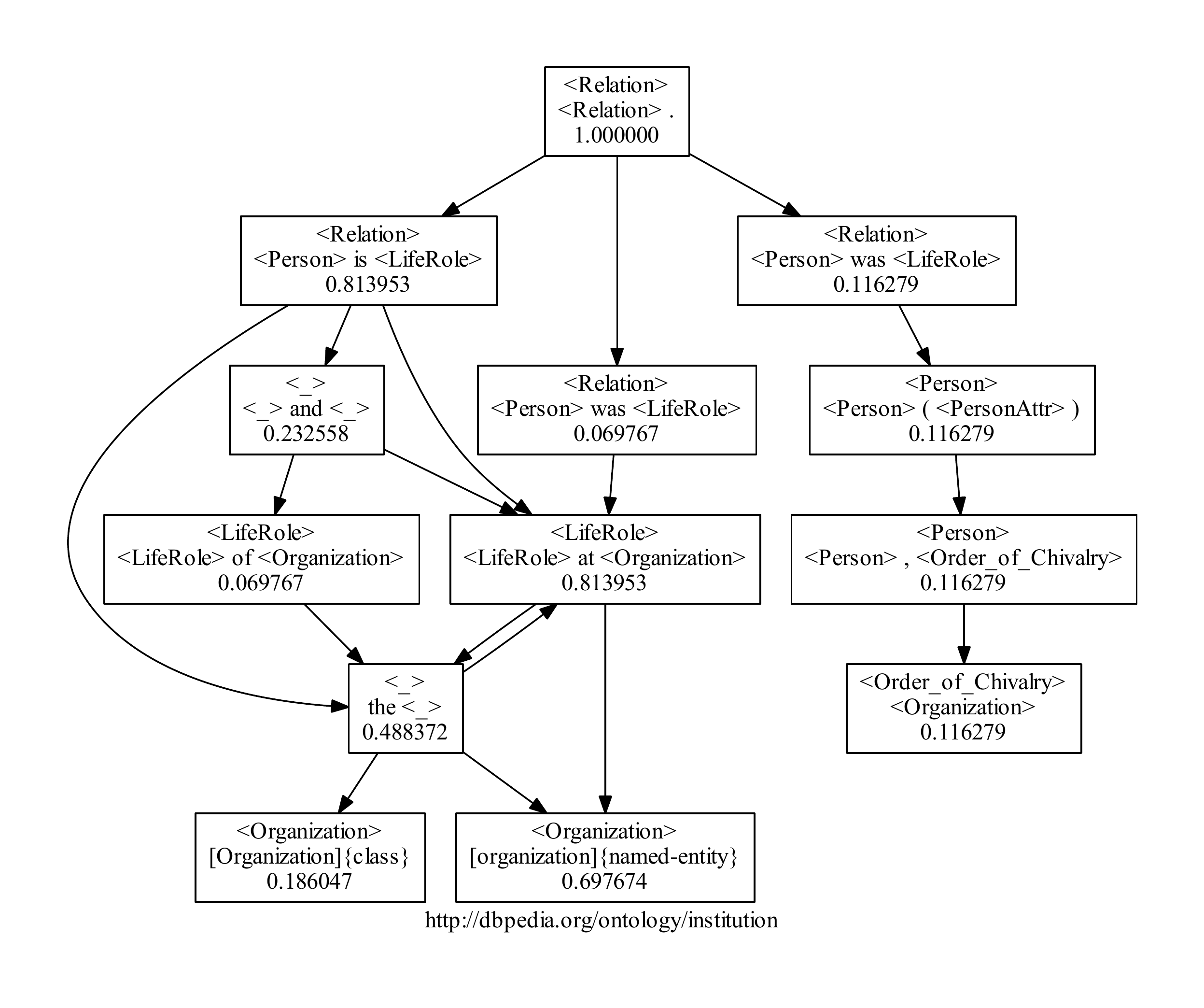}

			\caption{\small{The automaton model for \emph{institution} relation. On the bottom of each node, the fraction of training variable trees that contain this node, is displayed.}}
			\label{fig:net_model}
		\end{figure}
	\end{center}

	Several models based on variable trees were trained for solving this classification problem:
	\begin{itemize}
		\item \emph{Basic} (\emph{Basic model}) -- The model contains positive trained variable trees. In the prediction, if the test variable tree matches one of the trees in the model, then the example is predicted positive.
		\item \emph{Net} (\emph{Automaton model}) -- All positive variable trees are paths with start and end points. In this model they are merged into a net, which acts as a deterministic automaton. If the automaton accepts the test variable tree, than it is predicted positive. An example of automaton model is presented in Figure~\ref{fig:net_model}.
		\item \emph{LR} (\emph{Logistic regression}) -- A logistic regression model is trained with positive and negative examples, where nodes in variable trees represents features.
		\item \emph{LRC} (\emph{Logistic regression + Context nodes}) -- All leaf nodes that are siblings of any of the nodes in the variable tree are added to the \emph{LR} model.  
		\item \emph{LRCL} (\emph{Logistic regression + Context nodes + Lexical Tokens}) -- Tokens from the lexical layer of the entity nodes are added to the \emph{LRC} as features.

	\end{itemize}
	
	For training all or a maximum of 10.000 eligible relations was taken for each of 74 relation types. A 10-fold cross validation was performed for evaluation. The results are presented in Table~\ref{tab:f1}. The converted recall and converted F1 score presents recall and F1 on converted examples, which are the one, where relations were successfully converted into variable trees. The performance increases with each model, however the interpretability decreases. We also compared our method to the conditional random fields(CRF). In the CRF method, tokens from all layers with window size 7 were taken as features for sequence prediction. On the converted examples CRF achieved F1 score of 80.8, which is comparable to our best model's (LRCL) F1 score of 80.0.   
	
	\begin{table}
		\centering

		\begin{tabular}{llllll}
			
			\hline
			\textbf{Model} & \textbf{Precision} & \textbf{Converted Recall} &  \textbf{Recall} & \textbf{Converted F1} & \textbf{F1}\\
			
			\hline
			Basic & 72.2 &	48.3 &	28.2 & 52.1 & 36.4	\\
			Net & 61.6 & 77.6 &	42.5 &	63.3 &	 46.9\\
			LR & 71.9 &	84.2 & 50.2	 &	77.0 &	55.5 \\	
			LRC & 72.4 &	84.4 & 50.3 &	 77.3	 &	55.7\\
			LRCL & 76.2 &	85.5 &  50.7 & 80.0	& 57.1 \\	
			\hline
		\end{tabular}
		\caption{Performance of various relation extraction models.}
			\label{tab:f1}
	\end{table}

	\section{Related Work}
	\label{sec:related}
	
	There are many known approaches to ontology learning and semantic parsing, however, to the best of our knowledge, this is the first work to jointly learn an ontology and semantic parser. In the following sections, we make comparisons to other work on semantic parsing, ontology learning, grammar induction and others.

	\subsection{Semantic parsing}
	
	The goal of semantic parsing is to map text to meaning representations. Several approaches have used Combinatory categorial grammar (CCG) and lambda calculus as a meaning representation \cite{kwiatkowski2011lexical,krishnamurthy2015learning}. CCG grammar closely connects syntax and semantics with a lexicon, where each entry consist of a term, a syntactical category and a lambda statement. Similarly, our context-free grammar contains production rules. Some of these rules do not contain lexical tokens (the grammar is not lexicalized), which gives ability to express some relations with a single rule. For instance, to parse \emph{jazz drummer}, rule $<$Musician\_Type$>$ ::=$<$Musical\_Genre$>$ $<$Musician\_Type$>$ is used to directly express the relation, which determines the genre of the musician. Lambda calculus may provide a more formal meaning representation than semantic trees, but the lexicon of CCG requires mappings to lambda statements. Other approaches use dependency-based compositional semantics \cite{liang2013learning}, ungrounded graphs \cite{reddy2014large}, etc. as meaning representations.
	
	Early semantic parsers were trained on datasets, such as Geoquery \cite{zelle1996learning} and Atis \cite{Dahl:1994:ESA:1075812.1075823}, that map sentences to domain-specific databases. Later on datasets for question answering based on Freebase were created -- Free917 \cite{2013-acl-textual-schema-matching} and WebQuestions \cite{berant2013semantic} These datasets contain short questions from multiple domains, and since the meaning representations are formed of Freebase concepts, they allow reasoning over Freebase's ontology, which is much richer than databases in GeoQuery and Atis. All those datasets were constructed by either forming sentences given the meaning representation or vice-versa. Consequently, systems that were trained and evaluated on these datasets, might not work on sentences that cannot be represented by the underlying ontology. To overcome this limitation \cite{krishnamurthy2015learning} developed a open vocabulary semantic parser. Their approach uses a CCG parser on questions to from labmda statements, which besides Freebase vocabulary contain underspecified predicates. These lambda statements are together with answers -- Freebase entities -- used to learn a low-dimensional probabilistic database, which is then used to answer fill-in-the-blank natural language questions. In a very similar fashion, \cite{ckz:ACL15} defines underspecified entities, types and relations, when the corresponding concept does not exist in Freebase. 
	In contrast, the purpose of our method is to identify new concepts and ground them in the ontology.

	\subsection{Ontology Learning}
	
	%Many approaches work on ontology learning, where the goal is to extract knowledge on several layers -- terms, concepts, taxonomy, and relations. 
	Many ontology learning approaches address the same ontology components as our approach. However, their goal is to learn only the salient concepts for a particular domain, while our goal is to learn all the concepts (including instances, like particular organizations), so that they can be used in the meaning representation.   
	As survey by \cite{wong2012ontology} summarizes, the learning mechanisms are based either on statistics, linguistics, or logic. Our approach is unique because part of our ontology is constructed from the grammar. Many approaches use lexico-syntactic patterns for ontology learning. These are often based on dependency parses, like in \cite{Zouaq:2009:EGD:1638612.1639245,volker2007acquisition}. Our approach does not rely on linguistic preprocessing, which makes it suitable for non-standard texts and poorly resourced languages. Our approach also build patterns, however in form of grammar rules. Instead of lexico-syntactic patterns, which contain linguistic classes, our approach models semantic patterns, which contain semantic classes, like \emph{Person} and \emph{Color}. 
	These patterns are constructed in advance, which is sometimes difficult because the constructor is not always aware of all the phenomena that is expressed in the input text. Our approach allows to create a small number of seed patterns in advance, then explore other patterns through process of grammar learning. A similar bootstrapping semi-automatic approach to ontology learning was developed in \cite{liu2005semi}, where the user validates lexicalizations of a particular relation to learn new instances, and in \cite{brewster2002user}, where the user validates newly identified terms, while in our approach the user validates grammar rules to learn the composition of whole sentences.   
	A similar approach with combining DBpedia with Wikipedia for superised learning has been taken in \cite{walter2014atoll}, however their focus is more on lexicalization of relations and classes.
	
	\subsection{Grammar induction}
	Our goal was to develop a semi-automatic method that induces a grammar suitable for our scenario, in which an ontology is extracted, and text is parsed into semantic trees. A survey by \cite{d2011survey} compares several papers on grammar induction. According to their classification, our method falls into unsupervised, text-based (no negative examples of sentences) methods. Many such methods induce context-free grammars. However, their focus is more on learning syntactic structures rather than semantic. This is evident in evaluation strategies, where their parse trees are compared against golden parse trees in treebanks, like Penn treebank \cite{marcus1993building}, which are annotated according to syntactic policies. Furthermore, our grammar should not limited to a specific form, like for instance Chomsky normal form or Greibach normal form, instead it may contain arbitrary context-free rules.
	Several algorithms, like ours, employ the greedy strategy of grammar induction, where the grammar is updated with the best decision at each step. Whereas our method adds a rule after all sentences are parsed, The Incremental Parsing algorithm \cite{seginer2007fast} updates the grammar after each sentence. This is also done in ADIOS method \cite{solan2005unsupervised}, where it has been shown that order of sentences affects the grammar. Our method employs frequency analysis and human supervision to control the grammar construction, while others use Minimum Description Length principle \cite{watkinson2001psychologically}, clustering of sequences \cite{clark2001unsupervised}, or significance of word co-occurrences \cite{hanig2008unsuparse}.
	
	\subsection{Other Approaches}   
	
	Related work linking short terms to ontology concepts \cite{starcsemi} is designed similarly as our approach in terms of bootstrapping procedure to induce patterns. But instead of inducing context-free grammar production rules, suggestions for rewrite rules that transform text directly to ontology language are provided. 
	Another bootstrapping semi-automatic approach was developed for knowledge base population \cite{wolfe2015interactive}. The task of knowledge base population is concerned only with extracting instances and relations given the ontology. In our work we also extract the backbone of the ontology -- classes and taxonomic relations.
	Also, many other approaches focus only on one aspect of knowledge extraction, like taxonomy extraction \cite{Navigli:2011:GAI:2283696.2283715, P06-1101} or relation extraction \cite{Bunescu:2005:SPD:1220575.1220666, Zelenko:2003:KMR:944919.944964}. Combining these approaches can lead to cumbersome concept matching problems. This problem was also observed by \cite{Poon:2010:UOI:1858681.1858712}. Their system OntoUSP tries to overcome this by unsupervised inducing and populating a probabilistic grammar to solve question answering problem. However, the result are logical-form clusters connected in an \emph{isa} hierarchy, not grounded concepts, which are connected with an existing ontology.
	
	\section{Discussion}
	\label{sec:discussion} 
	We have presented an approach for joint ontology learning and semantic parsing. The approach was evaluated by building an ontology representing biographies of people. The first sentences of person Wikipedia pages and the combination of DBpedia and Freebase were used as a dataset. 
	This dataset was suitable for our approach, because the text is equipped with human tagged annotations, which are already linked to the ontology. In other cases a named entity disambiguation would be needed to obtain the annotations. The next trait of the dataset, that is suitable for our approach, is the homogeneous style of writing. Otherwise, if the style was more heterogeneous, the users would have to participate in more iterations to achieve the same level of coverage. The participation of the users may be seen a cost, but on the other hand it allows them to learn about the dataset without reading it all. The users does not learn so much about specific facts as they learn about the second order information, like what types of relations are expressed and their distribution.  
	
	Semantic trees offer a compact tree-structured meaning representation, which could be exploited for scenarios not covered by this paper, like relation type discovery and question answering. Furthermore, they can be used for more interpretable representation of meaning, like the automaton representation in Figure~\ref{fig:net_model}, compared to some other methods, like the one based on neural networks \cite{blunsom2014deep}. 
	Our approach may not be superior on one specific part of the ontology learning, but it rather provides an integrated approach for learning on several levels of the ontology.
	Also, our approach does not use syntactic analysis, like part of speech tags or dependency parsing, which makes our approach more language independent and useful for non-standard texts, where such analysis is not available. On the other hand, we are looking into integrating syntactic analysis for future work. One scenario is to automatically detect the property of the rule. Another idea for future work is to integrate some ideas from other grammar induction methods to detect meaningful patterns without relying on the annotation of text.

	\begin{acknowledgements}
		This work was supported by Slovenian Research Agency and the ICT Programme of the EC under XLike (FP7-ICT-288342-STREP) and XLime (FP7-ICT-611346).
	\end{acknowledgements}

\bibliography{paper} 

\end{document}